\documentclass[11pt,a4paper]{article}
\usepackage[hyperref]{acl2021}
\usepackage{times}

\usepackage{latexsym}
 \usepackage{graphicx}
 \usepackage{multirow}

\usepackage{microtype}

\aclfinalcopy 

\title{Exploring Code Style Transfer with Neural Networks}

\author{Karl Munson*, Anish Savla*, Chih-Kai Ting*, Serenity Wade* \\ University of California, Santa Cruz \\ \{kmunson, asavla, cting3, swade1\}@ucsc.edu
        \AND
 Kiran Kate \and Kavitha Srinivas \\ IBM Research}

\date{}

\begin{document}
\maketitle
\begin{abstract}
Style is a significant component of natural language text, reflecting a change in the tone of text while keeping the underlying information the same. Even though programming languages have strict syntax rules, they also have style. Code can be written with the same functionality but using different language features. However, programming style is difficult to quantify, and thus as part of this work, we define style attributes, specifically for Python. To build a definition of style, we utilized hierarchical clustering to capture a style definition without needing to specify transformations. In addition to defining style, we explore the capability of a pre-trained code language model to capture information about code style. To do this, we fine-tuned pre-trained code-language models and evaluated their performance in code style transfer tasks.
    \\
    
\end{abstract}

\section{Introduction}

\subsection{Motivation}
Recent years have seen a sharp increase in research on code language modeling, with work ranging from pre-training large language models over code (\citet{plbart}, \citet{codeT5}, \citet{codebert}), as well as related work examining the application of other natural language processing tasks applied to code, such as code machine translation or code generation. \citet{codemt}, \citet{codex}.  However, accompanying this rise has been a lack of work studying how well these models understand the human notion of code style. 

Code style is an integral element of code, so many organizations regulate code style through code reviews to improve readability and maintainability. Code style is, broadly, the appearance of code to a viewer and is thus - as much the same as writing style - something very individual to its writer. Due to the strict syntactic grammars governing how code is structured, many aspects of code style can be quantified automatically. \citet{PythonLanguage} programmatically studies the usage of Python language features across publicly available repositories. 

We assert that the content of code (its functionality) and the style of code (its aesthetics) are, like the content and style of text, aspects whose latent representations can be extricated and individually manipulated through neural style transfer. 

In this work, we defined an embedding space whose dimensions are extricable metrics measuring different aspects of code style. We then explored how to analyze code style through clustering in this space and how well we can classify code into these clusters. From there, we generated data and fine-tuned, pre-trained code language models to perform code style transfer along one aspect of code style.

\subsection{Contribution}
The main contributions of this work are fourfold. First, we applied clustering algorithms over code scripts embedded in a code style embedding space. We did this to find code clusters with distinct styles and evaluate the quality and predictability of the discovered clusters. Second, from publicly available code repositories, we generated a suite of parallel corpora; each corpus represents an atomic style transformation, such as casing or comments. Third, we used these parallel corpora to fine-tune pre-trained code language models on sequence-to-sequence individual style transfer tasks. These models can perform precise stylistic alterations one at a time, such as modifying casing or adding comments. We also trained a joint model that can apply multiple style transfers at once. Finally, whereas the existing workaround code language models focus on the syntactic integrity of processed code, our work focuses on something novel: code style.

\section{Related Work / Background}
\subsection{Code Style}
Due to the subjective nature of code style, the first background area of interest to address is what constitutes code style. This paper focuses primarily on style aspects of Python as a result of the large number of stylistic choices available in Python, large number of publicly available code samples, and familiarity with the language.

Many of the python language features that we focused on were selected based on the analysis done in An Empirical Study for Common Language Features Used in Python Projects \cite{PythonLanguage}. This paper tries to study the use and impact of Python language features in real-world Python projects. The authors analyze Python language features and automatically identify the use of 22 kinds of common Python language features in 6 categories in Python source code. This study was conducted over 35 popular Python projects from eight application domains covering 4.3 million lines of code; to investigate the usage of these language features in the project. Inheritance, decorator use, keyword arguments, for loops, and nested classes are listed as the top 5 used language features.

\subsection{Code Language Models}
Recent years have seen the natural language processing literature dominated by the pre-train and fine-tune paradigm. In line with this, we will primarily use two pre-trained language models for code for our experiments, PLBART and CodeT5. Pre-trained model usage will eliminate the need for extensive resources for training a code language model from scratch. 

\textbf{CodeT5} is an encoder-decoder model pre-trained on masked span prediction, identifier tagging, and masked identifier prediction for bimodal, programming, and natural language, dual generation \cite{codeT5}. It uses CodeSearchNet data which consists of unimodal (programming languages only) and bimodal (programming language and natural language) data on six programming languages. Developed by Salesforce, it is available on HuggingFace in different model sizes.

\textbf{PLBART} is another encoder-decoder model we explored. The model is pre-trained in Java, Python, and natural language on reconstruction and denoising \cite{plbart}. Like CodeT5, it is publicly available on HuggingFace in different model sizes, and has checkpoints fine-tuned on various code-related tasks.

Even though language models for code are used for many tasks, such as bug detection, code translation, and code generation; to the best of our knowledge, our work is the first one to address the task of code style transfer.
\\

\subsection{Text Style Transfer}
\begin{figure}[h!]
	\centering
	\includegraphics[width=\linewidth]{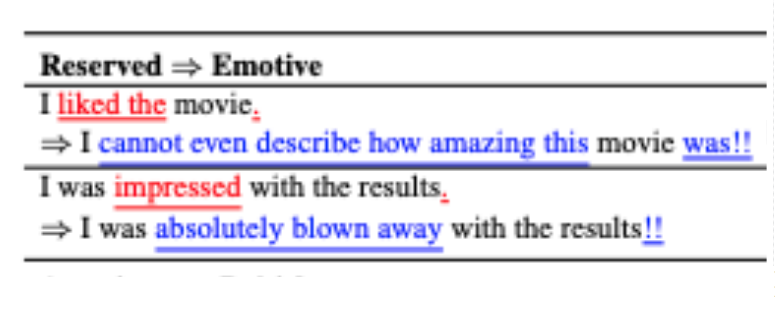}
	\caption{\label{fig:text_style_transfer} Example of text style transfer. \cite{riley2021textsettr}}
\end{figure}

Text style transfer is the task of reformatting a natural language text from one style to a text with the same meaning in a different style, while maintaining the textual content of the source.(See figure \ref{fig:text_style_transfer}) There are a number of ways to approach this task, and a significant amount of work has been done to use neural models for text style transfer (\citet{riley2021textsettr}, \citet{Tikhonov_2019}, \citet{zero_shot_text_style_transfer}, \citet{tst_survey}). This existing work in text style transfer in natural language serves as a blueprint on how to apply the same style transfer to code. 

\begin{figure}[h!]
	\centering
	\includegraphics[width=\linewidth]{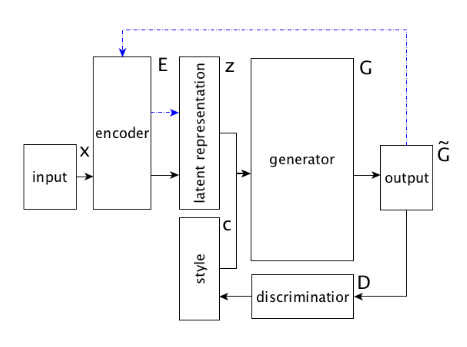}
	\caption{\label{fig:style_transfer_model} Example of GAN style transfer architecture. \cite{Tikhonov_2019}}
\end{figure}

Most approaches to this task utilize either an encoder-decoder framework or a generative adversarial network with different variations to those two basic frameworks, such as the one pictured below (\ref{fig:style_transfer_model}).

\section{Requirements / Proposal}
\subsection{Defining Code Styles}

The core underlying topic of this research is code style, which is, most broadly, the appearance of the code. Code style is vital to the readability of code. Many organizations value code styles which adhere to a centralized style guide and conduct code reviews to ensure that the style of written code matches a certain quality threshold. Though code style is subjective and very complex, we develop a framework for analyzing specific aspects of code style through quantifiable metrics and use this framework to conduct experiments with both clustering and sequence-to-sequence model training. 

Listed below are definitions of the aspects of code style we decided to focus on. These categories describe the main attributes of code style we chose to focus on for parallel corpora generation. We also used additional attributes during clustering experiments, mostly related to the above-described traits.

\subsubsection{Casing}
For this metric, we consider the various types of casing found within code: snake case, lower and upper camel case, and lowercase. Also, we consider how these casing types appear in different varieties of identifier names, such as variable, method, or class names. 

\subsubsection{Comment}
Comments are vital elements of code style as they greatly aid the readability of the code. We consider many attributes of code comments to be style-defining. Comment word length, comment density, and the ratio of commented lines in a script all act as approximate measures of how much a script has been annotated with comments.

\subsubsection{Docstrings}
Docstrings, like comments, are written by the programmer to aid the readability of the code and are integral to code style. They serve as headers to inform readers of the intended function, input and output arguments, and types of a method, class, or module. They often contain other information as well, relating to copyright or authorship.

\subsubsection{List Comprehensions}
We additionally focus on Python language-specific features, such as comprehensions. List comprehensions are one such language feature, used to iterate and operate on the items within a data structure. A programmer's decision between using a for loop or comprehension to do this is a Python-specific stylistic choice.

\subsubsection{Usage of Classes}
Being an object-oriented language, Python allows programmers to define classes. How one chooses to organize classes, design the methods within classes, or use inheritance, are all additional hallmarks of code style.

\subsection{Code Style Transfer}
Each code sample has a style as defined in the previous section. We want to be able to map a target style on to a sample of code which does not have that style. This is the same task as text style transfer but applied to code snippets.

\section{Design}
\subsection{Clustering}
Code style is difficult to define, so we selected several features (See Table \ref{code_style_feature}) to embed Python code into a multidimensional space. Then we used unsupervised clustering algorithms to identify the types of code that are most similar to each other. From these clusters, we define spatial buckets that represent different and distinct code styles. 

To cluster style features, we utilized the HDBSCAN\cite{hdbscan} clustering model to group code samples together based on similar counts of certain style features. HDBSCAN is a hierarchical clustering algorithm that we approximated by a random forest classifier to perform the prediction of cluster labels on new data. 
\begin{table*}[ht!]
\begin{center}

\begin{tabular}{ll}
\hline
\multicolumn{2}{c}{Code Style Features for Clustering}                        \\
\hline
Feature Type                     & Features                                   \\
\hline

\multirow{9}{*}{Casing}          & Snake case variable usage ratio            \\
                                 & Snake case function name usage ratio       \\
                                 & Snake case class usage ratio               \\
                                 & Upper camel case variable usage ratio      \\
                                 & Upper camel case function name usage ratio \\
                                 & Upper camel case class usage ratio         \\
                                 & Lower camel case variable usage ratio      \\
                                 & Lower camel case function name usage ratio \\
                                 & Lower camel case class usage ratio         \\
\hline
\multirow{2}{*}{Documentation}   & Docstring Density                          \\
                                 & Comment Density                            \\
\hline
\multirow{3}{*}{Function/Class}  & Average function decorators usage count    \\
                                 & Average class decorators usage count       \\
                                 & Average class inheritance usage count      \\
\hline
\multirow{3}{*}{Python Features} & Average list comprehension usage count     \\
                                 & Average generators usage count             \\
                                 & Average lambda function usage count        \\
\hline
\end{tabular}

\caption{The selected code style features which were used for clustering experiments.}
\label{code_style_feature}
\end{center}
\end{table*}

\subsection{Style Transfer GAN}
We adopt a similar style transfer architecture to \citeauthor{Tikhonov_2019}, the state-of-the-art text style transfer on the Yelp small review sentiment transfer task. The model will be slightly different from the original paper, so the pre-trained encoder-decoder is uninterrupted by inserting style between the components. The input code is concatenated with the style code and reduced using a linear layer so that it fits into the pre-trained encoder. The pre-trained encoder decoder will take this compressed input and produce a styled code sample that goes into the discriminator. We used the discriminator loss to modify the style representation concatenated to the input.
\paragraph{Substitute Pre-trained Model} We replace the encoder from \citeauthor{Tikhonov_2019} with the encoder of our pre-trained model CodeT5 or PLBART. We use the pre-trained encoder to generate a latent representation from the input code snippet. For the model generator, we will use the decoder of CodeT5 or PLBART. The generator will output decoded code. This code passes to the discriminator, a 3-layer multi-layer perceptron; again following from \citeauthor{Tikhonov_2019}. The MLP will learn to represent the style of the generated code, and a style loss passes back to the encoder.
\subsection{Style Classification with Pre-Trained Models}
The GAN-trained model did not generate intelligible Python code let alone correctly styled code. This begs the question, do pre-trained code language models even have a notion of code styles? \\ 

As a result, we built a classification model from the pre-trained language models we considered to see if the pre-trained models could capture any style information from the HBDSCAN clusters. The inputs for this model are code samples, and the outputs for the model are the labels generated from the clustering experiments. We implemented two versions of the classifier, one for PLBART and one for CodeT5. The PLBART model is a pre-trained PLBART model with a sequence classification head. The CodeT5 model uses a CodeT5 model and fine-tunes the generation model to produce the labels via generation instead of through a classification head.

\subsection{Seq2Seq Style Transfer}
The style classification results showed pre-trained code language models have some ability to capture style elements. As a result, we constructed a sequence-to-sequence model to do style transformations. Instead of transferring from one style to another based on complex features, we focused on individual style transfers like going from a for loop to an equivalent list comprehension. To do this, we constructed parallel corpora for five transfer tasks, generating comments, generating docstrings, adding casing, transforming code to a class structure, and converting for loops to list comprehensions when relevant. The details for parallel corpora generation are in the data section. 

The sequence-to-sequence model is a pre-trained CodeT5 small instance with 60 million parameters fine-tuned on the parallel corpora. For each of the individual transformations, we fine-tuned a sequence-to-sequence model to see if it was capable of performing specified single style transforms. In addition, we also fine-tuned a combined sequence-to-sequence model on all of the parallel corpora. The resulting multi-task model can do all the transformations with only one instance of CodeT5 instead of five. To fine-tune the model on multiple tasks, the dataloader randomly sample batches from each of the five parallel corpora. The dataloader samples proportionally to the size of the parallel corpora meaning that larger corpora are sampled more based on their size. 

The combined sequence-to-sequence model also had a modification to its training data. For each code sample, a natural language prompt, separated from the code by a special token, was used to indicate to the model which transformation the model should perform (figure:\ref{single_prompt}). This allows the desired transformation to be specified during the inference of the model and for prompting of multiple transformations to be performed on a single code sample. 

\section{Data}
\subsection{Py150k}
We used Py150k \cite{py150k}, which consists of parsed Abstract Syntax Trees from Python 2.7 scripts collected from a set of Github repositories with permissive and non-viral licenses. The dataset creators removed duplicate files and project forks, keeping only programs that parse and have at most 30,000 nodes in the AST. The set encompassed code from 5958 unique Github users and 8422 Github repositories. The dataset is split into two parts: 100,000 files for training and 50,000 for evaluation. 
\subsection{BigQuery}
We also used Google BigQuery for additional sources of Python code. The set contains 1.3 million python files drawn from Github with the watch count available. The data is split into two parts, samples from the CuBERT\cite{cubert} dataset and new code samples which were added to BigQuery after the CuBERT paper.

The files from the CuBERT dataset are already deduplicated to avoid contamination of dataset with py150, and duplicate scripts \cite{cubert}. For non-deduplicated files, we did the deduplication via their paths in the initial query to remove direct copies of code samples. In order to ensure deduplication of the remaining files,  we calcluated an MD5 checksum on the contents of each file, and removed duplicates from the dataset. The query used to collect the big query data was based on the one used in the CuBERT paper and will be made available in addition to other resources used in this research.
\\

\subsection{Parallel Corpora}
For the sequence-to-sequence transformation, we generated parallel corpora for each individual style transformation for training. For all transformations other than the comment transformation, the input and labels for code samples had no comments as the AST module used for transformations strips comments; so for the transformed data to match the original, the original code also needs comments removed. These parallel corpora transformations are easy to produce programmatically, but are difficult to undo. The model needs to identify when to make transformations and correctly produce them which is not a trivial solution. 

\subsubsection{Casing} For casing, we selected identifiers and method names using the AST module, lower-cased all of the data and stripped out underscores. This data with uncased code is the input, and the label is the code sample with the original casing. The model thus learns how to format identifier names. There are 700,000 examples in this dataset.
\subsubsection{Docstring} Code samples were split using the AST module isolating methods with corresponding docstrings. The input to the sequence-to-sequence model has the docstrings stripped; the label is the original code with the docstring. There are 2,500,000 examples in this dataset.
\subsubsection{Comments} We took code samples and parsed them, then unparsed the code using the AST module. This parse strips the comments and returns the same code with only comments removed. The input is the uncommented code; the label is the original commented code. There are 700,000 examples in this dataset.
\subsubsection{Classes}
We took code samples with classes and removed the class definitions and all references to self in methods. The input for this model is the code without class-related syntax, and the gold label is the original code. There are 400,000 examples in this dataset.
\subsubsection{List Comprehensions}
Code samples with list comprehensions in them are selected via the AST. To make the versions without list comprehensions, each list comprehension is converted to an equivalent function. The input for this model is the code with list comprehensions converted to for loops, and the gold label is the original code. This dataset has 36,000 examples.


\section{Experiments}
\subsection{Code Style Clustering}

We collected metrics(table \ref{code_style_feature}) for the scripts based on the style features specified in the design section and then used HDBScan \cite{hdbscan} to find and evaluate clusters of code. HBDSCAN has three parameters that we tuned to reduce the number of clusters while keeping clusters distinct.  \\

In addition to measuring cluster distinctness, we also measured the predictability of the cluster results using several non-neural classifiers. The experiment tested whether the features have enough signals to build a high-quality classification model since the code style could be sophisticated and difficult to characterize. We tested SVM, decision trees, random forest, naive Bayes, and logistic regression models' ability to predict cluster labels given the constructed style space for a script.

\subsection{Style Transfer GAN}
To test the viability of this model, we implemented it as described in the design section with two different configurations of pre-trained models. In one case, we used CodeT5; in the other, we used PLBART as the pre-trained component. In both cases, the GAN model could not output valid python code, let alone perform the style transfer task, despite tuning parameters and troubleshooting the model. As a result, we aborted the experiments with this architecture.
\subsection{Style Cluster Classification}
The failure of the Style Transfer GAN could indicate that neural network architectures could not capture the styles defined with our clusters. As a result of this possibility,  we tested if CodeT5 or PLBART were capable of capturing code style at all. In order to test this, we tested different classifiers built on pre-trained code language models. We tested the CodeT5 small, 60 million parameters, and base, 120 million parameters. For PLBART, we tested with the ``uclanlp/plbart-multi\_task-python" checkpoint on HuggingFace as well as ``uclanlp/plbart-base." Both PLBART models have the exact parameter count but with different fine-tuning. For the hyperparameters of both models, we used a learning rate of 1e-4, weight decay of 0.01, and trained for four epochs. We chose these parameters as similar pre-trained classifiers used these parameters. Four epochs of training were finished as validation loss was not improving much. 
The rest of the hyperparameters used are default HuggingFace parameters for each model.

\subsection{Seq2Seq Style Transfer}
Style cluster classification showed the ability of pre-trained models to capture some elements of style. CodeT5 had the best performance for the classification task, so we next trained a set of sequence-to-sequence models to perform each style transfer. All these experiments were based on CodeT5 small because the base version had marginally better performance. However, more parameters make it less efficient to train multiple sequence-to-sequence models. We built a baseline with CodeT5 architecture but trained it from scratch to evaluate how much CodeT5 can learn with versus without pre-training code.
\\
\subsubsection{Individual Style Transfer}
We fine-tuned CodeT5 for each parallel corpora we generated and trained all models with the same hyperparameters. Each model was fine-tuned for four epochs for docstring, comments, casing, and class corpora. List comprehensions corpora had only 36,000 examples, and the corresponding task was thus fine-tuned for six epochs. Docstring was fine-tuned on a 400,000 example subset of the 2.5 million examples to reduce training time and make dataset size comparable to other tasks. 
For the baseline of this model, we used a randomized weights version of CodeT5 to train on the task. This model would have same architecture but none of the information about code learned in pre-training. By doing this we could show what parts of code style are captured in pre-training.

\subsubsection{Multiple Style Transfer}
The combined model was trained with a subset of each parallel corpora. For docstring transfer, 250,000 examples were used; for casing 200,000 examples were used; for comments, 350,000 examples were used; for class, 350,000 examples were used; and for list comprehensions, 36,000 examples were used. Each data sample had a natural language prompt attached to the front of the input to indicate the transformation to be performed. 
The model samples batches proportionally to the size of training data, and each batch only contained examples from one parallel corpus. The model was trained for 12 epochs over this dataset with the same hyperparameters as the individual models. The model was tested every 4 epochs for performance on each of the individual tasks using the evaluation set. \\

In addition to evaluating individual tasks, the model was evaluated using the evaluation set on doing multiple transformations at once. The model was given a natural language prompt(figure \ref{single_prompt} and figure \ref{multi_prompt}) that specified that multiple transformations should be done. Examples from the evaluation set where multiple transformations could be done were selected. Several combinations of features were performed, with at most five features transferred at once. All of these multiple transformations are zero-shot tasks, with the idea being to test the model's ability to use what it learned in the individual tasks without creating different combinations of parallel corpora for fine-tuning multiple transformations.

\section{Evaluation}
We report our evaluation on multiple components: clustering to define code style and style transfer using sequence to sequence generation. 

\subsection{Code Style Clustering}
For clustering, we evaluated the internal and external validity of the clusters. 

\subsubsection{Metrics}
\paragraph{Davies-Boules Index}
For the internal metric, the Davies-Boules Index(DBI)\citep{dbi} was selected; DBI measures the density of a cluster by calculating similarity across its containing data points. The score is considered better for the smaller value of DBI.

\paragraph{Purity}
For the external metric, we used Purity\citep{purity} to measure how well the clusters mapped to authorship of code scripts, with the assumption that a given author or a given organization will have the same style. Purity evaluates if a cluster can be recognized as a single class. 

\subsubsection{Qualitative Analysis}

\paragraph{Hyperparameter Tuning} 
\begin{table*}[ht!]
\begin{center}
\begin{tabular}{l|rrr|rr}
\hline
\multicolumn{5}{c}{Evaluation and Setup of our clusterer} \\
\hline 
Model & \multicolumn{3}{c|}{Hyperparameters}                                 & \multicolumn{2}{c}{Evaluation} \\
\hline
  & Min Cluster Size & Min Samples & Cluster Epsilon & Purity          & DBI          \\
\hline
Baseline               & -               & -           & -               & 0.246               & -        \\
Our clusterer          & 400             & 100         & 0.01            & \textbf{0.404}      & 1.06     \\
\hline
\end{tabular}
\caption{The performance and setup for the final selected clusterer.}
\label{cluster_score}
\end{center}
\end{table*}
For the results of the tuning experiments of HDBScan, we tuned three hyperparameters, which are the minimum of each cluster size(Min Cluster Size), minimum samples for calculating the distance to the centroid(Min Samples), and Cluster Epsilon for the vague of the boundary between each cluster. The best setup gave the best performance on both Purity and DBI. For Purity, the table \ref{cluster_score} shows that our clustering model performed a significant improvement from the random sampling baseline, randomly assigning data points to clusters and calculating the Purity with author labels. \\

\begin{figure}[ht!]
	\centering
	\includegraphics[width=\linewidth]{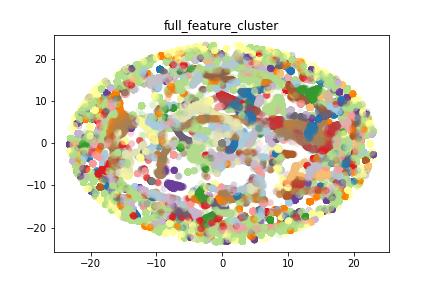}
	\caption{t-SNE\citep{tsne} visualization of the final clusters}
	\label{cluster_fig}
\end{figure}

Given the selected hyperparameters, we generated the final clusters from the entire dataset. Figure \ref{cluster_fig} shows the t-SNE visualization of the data metrics with colors representing the cluster of the data point. The light green points indicate the data detected as outliers, around 741K examples. Other than those, around 234K examples were clustered out of 975K.  \\

\paragraph{Pair Authorship Analysis} 
\begin{table*}[ht!]
\begin{center}
\begin{tabular}{l l|r r r r}
\hline
\multicolumn{6}{c}{Cluster Authorship Pair Comparison} \\
\hline
Authorship 1 & Authorship 2 & Purity & DBI & Purity-Improve & DBI-Improve \\
\hline
anhstudios & plotly    & \textbf{0.945} & 0.845          & \textbf{0.232} & \textbf{579.528} \\
anhstudios & AppScale  & 0.764          & \textbf{0.838} & 0.036          & 75.124           \\
anhstudios & dropbox   & 0.884          & 1.125          & 0.029          & 88.904           \\
anhstudios & mozilla   & 0.859          & 1.003          & 0.020          & 17.735           \\
anhstudios & google    & 0.767          & 1.082          & 0.015          & 32.572           \\
anhstudios & openstack & 0.635          & 0.881          & 0.015          & 94.766           \\
\hline
\end{tabular}
\caption{The comparison on authorship pair for the quality of the clusters.}
\label{author_compare}
\end{center}
\end{table*}
Furthermore, we compared the pairwise authorship across all clusters to evaluate the most distinctive coding style when assuming each author represents a style. The score will only consider the data points that belong to the pair of authors(authorship $1$ and authorship $2$). The improved score indicates how much it is improved from the random sampling baseline, randomly picking data points and assigning them to the clusters. The baseline score can be high simply because of the high authorship distribution. So the improvement score is considered for targeting the most valuable comparison by the HDBScan results. The table \ref{author_compare} shows the top comparing evaluation results. It suggests the \textit{anhstudios} and \textit{plotly} is having the most distinctive code style from others. Figure \ref{style_example_fig} shows an example of how their coding styles are different from each other.  \\

\paragraph{Traditional Machine Learning models}
\begin{table*}[ht!]
\begin{center}
\begin{tabular}{lrrrr}
\hline
\multicolumn{5}{c}{Performance on Traditional ML Classifier}                                     \\
\hline
Classifier    & Accuracy & Precision & Recall & F1 \\
\hline
SVM                    & 0.777    & 0.796    & 0.777    & 0.784    \\
Logistic Regression    & 0.736    & 0.798    & 0.736    & 0.752    \\
Decision Tree          & 0.980    & 0.980    & 0.980    & 0.980    \\
Random Forest & \textbf{0.989}               & \textbf{0.989}                & \textbf{0.989}             & \textbf{0.989}         \\
Naive Bayes            & 0.755    & 0.571    & 0.755    & 0.650 \\
\hline
\end{tabular}
\caption{This is the evaluation for testing how well the quality of clusters is by fitting the cluster labels and data metrics with various traditional machine learning classifiers.}
\label{ml_cluster}
\end{center}
\end{table*}
We also evaluated the clusters using a traditional classification algorithm with F-1 score, precision, recall, and accuracy. This evaluation tested the quality of clusters based on whether simple classifiers can map data points based on the same features we used for clustering into proposed clusters. Table \ref{ml_cluster} shows the performance score for measuring the predictability of the cluster for five different classifiers. The Naive Bayes was not performing well, randomly guessing the data points to the clusters. The logistic regression and SVM classifiers did relatively better but failed to recognize many cases. However, the tree-based classifiers performed exceptionally well, especially for the random forest classifier, which reached almost 99\% score for each metric. The reason depends on how the HDBScan splits the data points using hierarchical tree parsing. The results indicate that our clustering results perform good predictability on simple classifiers, but how will neural-based models do even without directly looking at the data features on which the clusters are based?

\paragraph{Pre-trained models Cluster Classification}
\begin{table*}[ht!]
\begin{center}
\begin{tabular}{lrrrr}
\hline
\multicolumn{5}{c}{Performance for Pre-trained Code Language Model}                                     \\
\hline
Model    & Accuracy & Precision & Recall & F1 \\
\hline
PLBart-Base              & 0.8           & 0.8           & 0.8           & 0.8           \\
PLBart-Multi-task-Python & 0.79          & 0.79          & 0.79          & 0.79          \\
CodeT5-Small             & 0.919 & \textbf{0.92} & \textbf{0.92} & \textbf{0.92} \\
CodeT5-Base              & \textbf{0.92} & \textbf{0.92} & \textbf{0.92} & \textbf{0.92} \\
\hline
\end{tabular}
\caption{The classification report for the pre-trained language model fine-tuned on cluster classification.}
\label{cluster_class}
\end{center}
\end{table*}
To ensure that pre-trained models could identify aspects of code style, we used F-1, precision, recall, and accuracy to evaluate if the pre-trained code language models, PLBart and CodeT5, could classify the clusters based purely on an input of source code. The table \ref{cluster_class} shows the classification performance of different checkpoints on these two architectures. CodeT5 performed a lot better results than PLBART on classifying the style clusters. The small checkpoint and the base checkpoint only had a very marginal difference in their scores, which means the compact version of CodeT5 was enough to capture the code style we defined with clusters. The better scores could contribute to the pre-training methods of CodeT5, which target token-level prediction, especially for identifiers. Identifiers defined by users contain much more style information about the code than other keyword tokens. It also has the pre-training task of bi-modal dual generation, which will understand the alignment between the code and the documentation. Overall, CodeT5 can capture the code style according to our proposed clusters even without seeing style features. However, will CodeT5 be able to perform style transfer?


\subsection{Code style transfer using Seq2Seq Generation}
We used the following metrics to evaluate sequence to sequence generation.
\subsubsection{Metrics}
\paragraph{CodeBLEU}
The sequence-to-sequence generation task is evaluated with CodeBLEU\citep{codebleu}, which is considered better than BLEU for code because it also compares code syntax and semantics. It leverages keywords of programming languages, the abstract syntax tree structure, and the semantics of programs. The final CodeBLEU score is an average value over BLEU score (N-gram), weighted BLEU score (Weighted N-gram), the syntax score (Syntax Match), and the semantic score (Dataflow Match).

\paragraph{BLEU}
For natural language tasks such as docstring transfer and comment transfer, evaluating them with the standard BLEU score\citep{bleu} is more appropriate. 

\paragraph{DiffBLEU}
Since the CodeBLEU is too optimistic when evaluating the transfer task, we proposed DiffBLEU, which is BLEU only on the difference between reference code $X$ and prediction code $\hat{Y}$ and the difference between input code $X$ and label code $Y$. CodeBLEU sometimes shows a high score which simply contributes to the successfully reconstructed code from the input other than the part that should be transferred. Especially for list comprehension transfer, the portion that needs to be modified by the model is significantly small. The CodeBLEU can still show high performance even without attempting any transfer. DiffBLEU can evaluate only the difference in the code that should be performed in the transfer task and applies to all transfer tasks. For natural language tasks such as docstring transfer and comment transfer, evaluating them with the standard BLEU score\citep{bleu} is more appropriate. 
\begin{equation}
    \textit{DiffBLEU} = \textit{BLEU}(\textit{Diff}(X, \hat{Y}), \textit{Diff}(X, Y))
\label{diffbleu_equation}
\end{equation}
The equation \ref{diffbleu_equation} shows the concept of the DiffBLEU calculation. The method of extracting difference, \textit{Diff}, is using the ``SequenceMatcher" function from the built-in library ``difflib". However, the DiffBLEU is not precise in capturing the difference when multiple positions and a large portion of change happen. DiffBLEU can only serve as a proxy for evaluating relative performance on the transfer part and will need a more precise metric, such as a parsed-tree-based difference score.


\paragraph{Parsability}
We have also evaluated the validity of the generated code with the parsability metric. Parsability is the accuracy of whether the code is still parsable from the AST module in Python3. Any error occurred during parsing will be considered as failing to parse.

\subsubsection{Individual Style Transfer}
\begin{table*}[ht!]
\begin{center}

\begin{tabular}{lrrrr}
\hline
\multicolumn{5}{c}{Overall Performance for Seq2Seq Generation on Individual Style Transfer}     \\
\hline
Task/Model & CodeBLEU & BLEU-NL & DiffBLEU & Parsability \\
\hline
\multicolumn{5}{l}{Comment Transfer}                                                    \\
\hline
Baseline             & 0.031          & 0             & 2.20E-82       & 0.000          \\
Individual Fine-tuned & \textbf{0.792} & \textbf{0.14} & \textbf{0.283} & \textbf{0.91}  \\
Combined Fine-tuned   & 0.779          & 0.049         & 0.23           & 0.896          \\
\hline
\multicolumn{5}{l}{Class Transfer}                                                      \\
\hline
Baseline             & 0.028          & -             & 1.52E-234      & 0.000          \\
Individual Fine-tuned & \textbf{0.848} & -             & \textbf{0.45}  & \textbf{0.98}  \\
Combined Fine-tuned   & 0.807          & -             & 0.37           & 0.918          \\
\hline
\multicolumn{5}{l}{Docstring Transfer}                                                  \\
\hline
Baseline             & 0.027          & 0             & 3.43E-234      & 0.000          \\
Individual Fine-tuned & 0.656          & 0             & \textbf{0.008} & \textbf{0.847} \\
Combined Fine-tuned   & \textbf{0.715} & 0             & 0.002          & 0.843          \\
\hline
\multicolumn{5}{l}{Casing Transfer}                                                     \\
\hline
Baseline             & 0.026          & -             & 2.53E-235      & 0.001          \\
Individual Fine-tuned & \textbf{0.967} & -             & \textbf{0.728} & \textbf{0.947} \\
Combined Fine-tuned   & 0.894          & -             & 0.382          & 0.925          \\
\hline
\multicolumn{5}{l}{List Comprehension Transfer}                                         \\
\hline
Baseline             & 0.021          & -             & 0.000           & 0.000          \\
Individual Fine-tuned & \textbf{0.982} & -             & \textbf{0.862} & \textbf{0.869} \\
Combined Fine-tuned   & 0.924          & -             & 0.499          & 0.793         	\\
\hline
\end{tabular}

\caption{The overall performance table on individual feature transfer.}
\label{indi_transfer_score}
\end{center}
\end{table*}
According to the table \ref{indi_transfer_score}, the baseline was reasonably performing scores close to zero since it is hard to transfer style without any pre-trained understanding of the programming language. The scores show that almost all individual fine-tuned models were performing the best to maintain generated code quality and perform transfer tasks.(See figure \ref{ind_example_fig} for example.) The combined fine-tuned model was trained with multi-task learning, giving noises affecting the individual transfer results. \\

However, for the docstring transfer, the combined model got better performance on CodeBLEU. One of the analyses is the combined model learned from a greater variety of training data than just the parallel corpus on docstring transfer. So its ability of code reconstruction can perform on a more extensive distribution of code compared to the individual model.

\paragraph{Manual Evaluation on Docstring Transfer}

\begin{table*}[ht!]
\begin{center}

\begin{tabular}{lrrrr}
\hline
\multicolumn{5}{c}{Manual Evaluation on Docstring Transfer}                                           \\
\hline
             				 	& \multicolumn{2}{c}{Sensible} & \multicolumn{2}{c}{Meaning} \\
\hline
Task/Model   	& Avg.        & Agreement      & Avg.       & Agreement      \\
\hline
\multicolumn{5}{l}{Docstring Transfer}                                                                \\
\hline
Individual Fine-tuned & 0.848       & 0.848          & 1.283      & 0.435    \\     
\hline
\end{tabular}

\caption{The manual evaluation on sensible and meaning of docstrings. The agreement score represent the percentage of how two annotators agree to each other.}
\label{man_eval_docstring}
\end{center}
\end{table*}
The manual evaluation was necessary for some natural language generation tasks such as docstring/comments transfer. Automatic scores for the natural language can mislead the true performance by giving a low score without considering the meaning of the generated text and the Sensibility of the model. So we evaluated the docstring transfer with manual evaluation on two metrics: \textbf{Sensibility} and \textbf{Meaning}.

\textbf{Meaning} metric was a $1$ to $3$ scale depending on how well the meaning of the predicted docstrings matched the reference docstrings. The table \ref{meaning_scale_table} shows the description of each scale. If a docstring did not have a corresponding docstring in the ground truth, it was considered as not having a similar meaning.

\textbf{Sensibility} was measured on a $0$ or $1$ basis to see if the docstrings produced for each code snippet are grammatical and not logical nonsense.

The manual evaluation(table \ref{man_eval_docstring}) indicates that the model can generate docstrings by sensing the contents of the code. However, the produced meanings are too simple and do not match well with the targets.

\subsubsection{Multiple Style Transfer}
\begin{table*}[ht!]
\begin{center}

\begin{tabular}{lrrrrr}
\hline
\multicolumn{6}{c}{Overall Performance for Seq2Seq Generation on Multiple Style Transfer}               \\
\hline
Task/Model & CodeBLEU & BLEU-Comment & BLEU-Docstring & DiffBLEU & Parsability \\
\hline
\multicolumn{6}{l}{Comment + Docstring}                                                                 \\
\hline
Baseline           & 0.035          & 0              & 0              & 4.53E-82       & 0              \\
Combined Fine-tuned & \textbf{0.560} & \textbf{0.003} & \textbf{0.005} & \textbf{0.003} & \textbf{0.839} \\
\hline
\multicolumn{6}{l}{Casing + Class}                                                                      \\
\hline
Baseline           & 0.026          & -              & -              & 1.51E-234      & 0              \\
Combined Fine-tuned & \textbf{0.669} & -    			 & -    		  & \textbf{0.097} & \textbf{0.894} \\
\hline
\multicolumn{6}{l}{List Comp + Casing + Class}                                                          \\
\hline
Baseline           & 0.019          & -              & -              & 0              & 0              \\
Combined Fine-tuned & \textbf{0.667} & -		         & -		      & \textbf{0.065} & \textbf{0.827} \\
\hline
\multicolumn{6}{l}{List Comp + Casing + Class + Docstring + Comment}                                    \\
\hline
Baseline           & 0.025          & 0              & 0              & 1.84E-157      & 0              \\
Combined Fine-tuned & \textbf{0.418} & \textbf{0.022} & \textbf{0.002} & \textbf{0.002} & \textbf{0.929} \\
\hline
\end{tabular}

\caption{The overall performance table on multiple feature transfer.}
\label{multi_transfer_score}
\end{center}
\end{table*}
According to the table \ref{multi_transfer_score}, the combined model performed significantly better than the baseline. However, the scores are performing in a lot worse range compared to the individual transfer. Even though the model can still generate the grammatical output, many predictions are the same as the input, not performing any transfer. Some only performed a subset of the transfer instead of the full combined transfer. However, there were also some successful cases(see figure \ref{mul_example_fig}) in which a full combined transfer was attempted. The observation indicates that the CodeT5-small revealed the capability of multiple code style transfers at once.



\section{Conclusion}
In this work, we performed a novel task - code style transfer with individual and multiple styles. We defined code styles through clustering and found that the style clusters reached decent quality and predictability. We explored pre-trained code language models, CodeT5 and PLBART, and found CodeT5 has the solid capability for capturing code style. We generated parallel corpora for our proposed tasks and fine-tuned CodeT5. The results strongly suggest that such a model can successfully learn the notion of code style and perform multiple style transfers at once.

\paragraph{Future Work} For further experiments, GAN training with the combined fine-tuned model as the generator will be an interesting test. Moreover, we can explore more powerful pre-trained language models such as BLOOM\footnote{https://huggingface.co/bigscience/bloom} and Codex\citep{codex} with prompting to examine whether and how they recognize the code styles.

\bibliographystyle{acl_natbib}
\bibliography{acl2021}

\begin{thebibliography}{18}
\expandafter\ifx\csname natexlab\endcsname\relax\def\natexlab#1{#1}\fi

\bibitem[{Ahmad et~al.(2021)Ahmad, Chakraborty, Ray, and Chang}]{plbart}
Wasi~Uddin Ahmad, Saikat Chakraborty, Baishakhi Ray, and Kai{-}Wei Chang. 2021.
\newblock \href {http://arxiv.org/abs/2103.06333} {Unified pre-training for
  program understanding and generation}.
\newblock \emph{CoRR}, abs/2103.06333.

\bibitem[{Chen et~al.(2021)Chen, Tworek, Jun, Yuan, de~Oliveira~Pinto, Kaplan,
  Edwards, Burda, Joseph, Brockman, Ray, Puri, Krueger, Petrov, Khlaaf, Sastry,
  Mishkin, Chan, Gray, Ryder, Pavlov, Power, Kaiser, Bavarian, Winter, Tillet,
  Such, Cummings, Plappert, Chantzis, Barnes, Herbert{-}Voss, Guss, Nichol,
  Paino, Tezak, Tang, Babuschkin, Balaji, Jain, Saunders, Hesse, Carr, Leike,
  Achiam, Misra, Morikawa, Radford, Knight, Brundage, Murati, Mayer, Welinder,
  McGrew, Amodei, McCandlish, Sutskever, and Zaremba}]{codex}
Mark Chen, Jerry Tworek, Heewoo Jun, Qiming Yuan, Henrique~Ponde
  de~Oliveira~Pinto, Jared Kaplan, Harrison Edwards, Yuri Burda, Nicholas
  Joseph, Greg Brockman, Alex Ray, Raul Puri, Gretchen Krueger, Michael Petrov,
  Heidy Khlaaf, Girish Sastry, Pamela Mishkin, Brooke Chan, Scott Gray, Nick
  Ryder, Mikhail Pavlov, Alethea Power, Lukasz Kaiser, Mohammad Bavarian,
  Clemens Winter, Philippe Tillet, Felipe~Petroski Such, Dave Cummings,
  Matthias Plappert, Fotios Chantzis, Elizabeth Barnes, Ariel Herbert{-}Voss,
  William~Hebgen Guss, Alex Nichol, Alex Paino, Nikolas Tezak, Jie Tang, Igor
  Babuschkin, Suchir Balaji, Shantanu Jain, William Saunders, Christopher
  Hesse, Andrew~N. Carr, Jan Leike, Joshua Achiam, Vedant Misra, Evan Morikawa,
  Alec Radford, Matthew Knight, Miles Brundage, Mira Murati, Katie Mayer, Peter
  Welinder, Bob McGrew, Dario Amodei, Sam McCandlish, Ilya Sutskever, and
  Wojciech Zaremba. 2021.
\newblock \href {http://arxiv.org/abs/2107.03374} {Evaluating large language
  models trained on code}.
\newblock \emph{CoRR}, abs/2107.03374.

\bibitem[{Davies and Bouldin(1979)}]{dbi}
David~L Davies and Donald~W Bouldin. 1979.
\newblock A cluster separation measure.
\newblock \emph{IEEE transactions on pattern analysis and machine
  intelligence}, (2):224--227.

\bibitem[{Feng et~al.(2020)Feng, Guo, Tang, Duan, Feng, Gong, Shou, Qin, Liu,
  Jiang, and Zhou}]{codebert}
Zhangyin Feng, Daya Guo, Duyu Tang, Nan Duan, Xiaocheng Feng, Ming Gong, Linjun
  Shou, Bing Qin, Ting Liu, Daxin Jiang, and Ming Zhou. 2020.
\newblock \href {https://doi.org/10.48550/ARXIV.2002.08155} {Codebert: A
  pre-trained model for programming and natural languages}.

\bibitem[{Jin et~al.(2020)Jin, Jin, Hu, Vechtomova, and Mihalcea}]{tst_survey}
Di~Jin, Zhijing Jin, Zhiting Hu, Olga Vechtomova, and Rada Mihalcea. 2020.
\newblock \href {https://doi.org/10.48550/ARXIV.2011.00416} {Deep learning for
  text style transfer: A survey}.

\bibitem[{Kanade et~al.(2020)Kanade, Maniatis, Balakrishnan, and Shi}]{cubert}
Aditya Kanade, Petros Maniatis, Gogul Balakrishnan, and Kensen Shi. 2020.
\newblock Learning and evaluating contextual embedding of source code.
\newblock In \emph{International Conference on Machine Learning}, pages
  5110--5121. PMLR.

\bibitem[{Van~der Maaten and Hinton(2008)}]{tsne}
Laurens Van~der Maaten and Geoffrey Hinton. 2008.
\newblock Visualizing data using t-sne.
\newblock \emph{Journal of machine learning research}, 9(11).

\bibitem[{Malzer and Baum(2020)}]{hdbscan}
Claudia Malzer and Marcus Baum. 2020.
\newblock A hybrid approach to hierarchical density-based cluster selection.
\newblock In \emph{2020 IEEE International Conference on Multisensor Fusion and
  Integration for Intelligent Systems (MFI)}, pages 223--228. IEEE.

\bibitem[{Manning(2008)}]{purity}
Christopher~D Manning. 2008.
\newblock \emph{Introduction to information retrieval}.
\newblock Syngress Publishing,.

\bibitem[{Papineni et~al.(2002)Papineni, Roukos, Ward, and Zhu}]{bleu}
Kishore Papineni, Salim Roukos, Todd Ward, and Wei-Jing Zhu. 2002.
\newblock \href {https://doi.org/10.3115/1073083.1073135} {{B}leu: a method for
  automatic evaluation of machine translation}.
\newblock In \emph{Proceedings of the 40th Annual Meeting of the Association
  for Computational Linguistics}, pages 311--318, Philadelphia, Pennsylvania,
  USA. Association for Computational Linguistics.

\bibitem[{Raychev et~al.(2016)Raychev, Bielik, and Vechev}]{py150k}
Veselin Raychev, Pavol Bielik, and Martin Vechev. 2016.
\newblock \href {https://doi.org/10.1145/3022671.2984041} {Probabilistic model
  for code with decision trees}.
\newblock \emph{SIGPLAN Not.}, 51(10):731–747.

\bibitem[{Reif et~al.(2021)Reif, Ippolito, Yuan, Coenen, Callison-Burch, and
  Wei}]{zero_shot_text_style_transfer}
Emily Reif, Daphne Ippolito, Ann Yuan, Andy Coenen, Chris Callison-Burch, and
  Jason Wei. 2021.
\newblock \href {https://doi.org/10.48550/ARXIV.2109.03910} {A recipe for
  arbitrary text style transfer with large language models}.

\bibitem[{Ren et~al.(2020)Ren, Guo, Lu, Zhou, Liu, Tang, Sundaresan, Zhou,
  Blanco, and Ma}]{codebleu}
Shuo Ren, Daya Guo, Shuai Lu, Long Zhou, Shujie Liu, Duyu Tang, Neel
  Sundaresan, Ming Zhou, Ambrosio Blanco, and Shuai Ma. 2020.
\newblock Codebleu: a method for automatic evaluation of code synthesis.
\newblock \emph{arXiv preprint arXiv:2009.10297}.

\bibitem[{Riley et~al.(2021)Riley, Constant, Guo, Kumar, Uthus, and
  Parekh}]{riley2021textsettr}
Parker Riley, Noah Constant, Mandy Guo, Girish Kumar, David Uthus, and Zarana
  Parekh. 2021.
\newblock \href {https://openreview.net/forum?id=T6RYeudzf1}
  {Text{\{}settr{\}}: Label-free text style extraction and tunable targeted
  restyling}.

\bibitem[{Szafraniec et~al.(2022)Szafraniec, Roziere, Leather, Charton,
  Labatut, and Synnaeve}]{codemt}
Marc Szafraniec, Baptiste Roziere, Hugh Leather, Francois Charton, Patrick
  Labatut, and Gabriel Synnaeve. 2022.
\newblock \href {https://doi.org/10.48550/ARXIV.2207.03578} {Code translation
  with compiler representations}.

\bibitem[{Tikhonov et~al.(2019)Tikhonov, Shibaev, Nagaev, Nugmanova, and
  Yamshchikov}]{Tikhonov_2019}
Alexey Tikhonov, Viacheslav Shibaev, Aleksander Nagaev, Aigul Nugmanova, and
  Ivan~P. Yamshchikov. 2019.
\newblock \href {https://doi.org/10.18653/v1/d19-1406} {Style transfer for
  texts: Retrain, report errors, compare with rewrites}.
\newblock In \emph{Proceedings of the 2019 Conference on Empirical Methods in
  Natural Language Processing and the 9th International Joint Conference on
  Natural Language Processing ({EMNLP}-{IJCNLP})}. Association for
  Computational Linguistics.

\bibitem[{Wang et~al.(2021)Wang, Wang, Joty, and Hoi}]{codeT5}
Yue Wang, Weishi Wang, Shafiq Joty, and Steven C.~H. Hoi. 2021.
\newblock \href {https://doi.org/10.48550/ARXIV.2109.00859} {Codet5:
  Identifier-aware unified pre-trained encoder-decoder models for code
  understanding and generation}.

\bibitem[{Yun~Peng(2021)}]{PythonLanguage}
Mingzhe~Hu Yun~Peng, Yu~Zhang. 2021.
\newblock \href {https://doi.org/10.48550/ARXIV.1909.09436} {An empirical study
  for common language features used in python projects}.

\end{thebibliography}

\appendix
\begin{figure*}[h!]
	\centering
	\includegraphics[width=\linewidth]{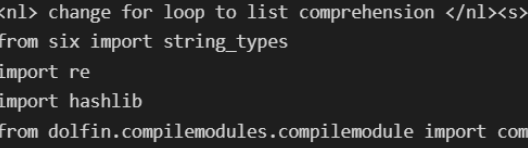}
	\caption{Example individual transfer prompt}
    \label{single_prompt}
\end{figure*}

\begin{figure*}[h!]
	\centering
	\includegraphics[width=\linewidth]{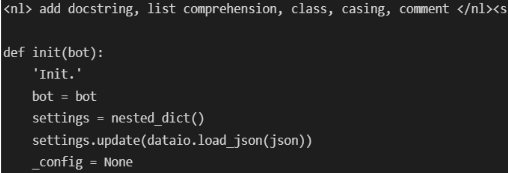}
	\caption{Example multi-transfer prompt}
    \label{multi_prompt}
\end{figure*}

\begin{figure*}[ht!]
	\centering
	\includegraphics[width=1\textwidth]{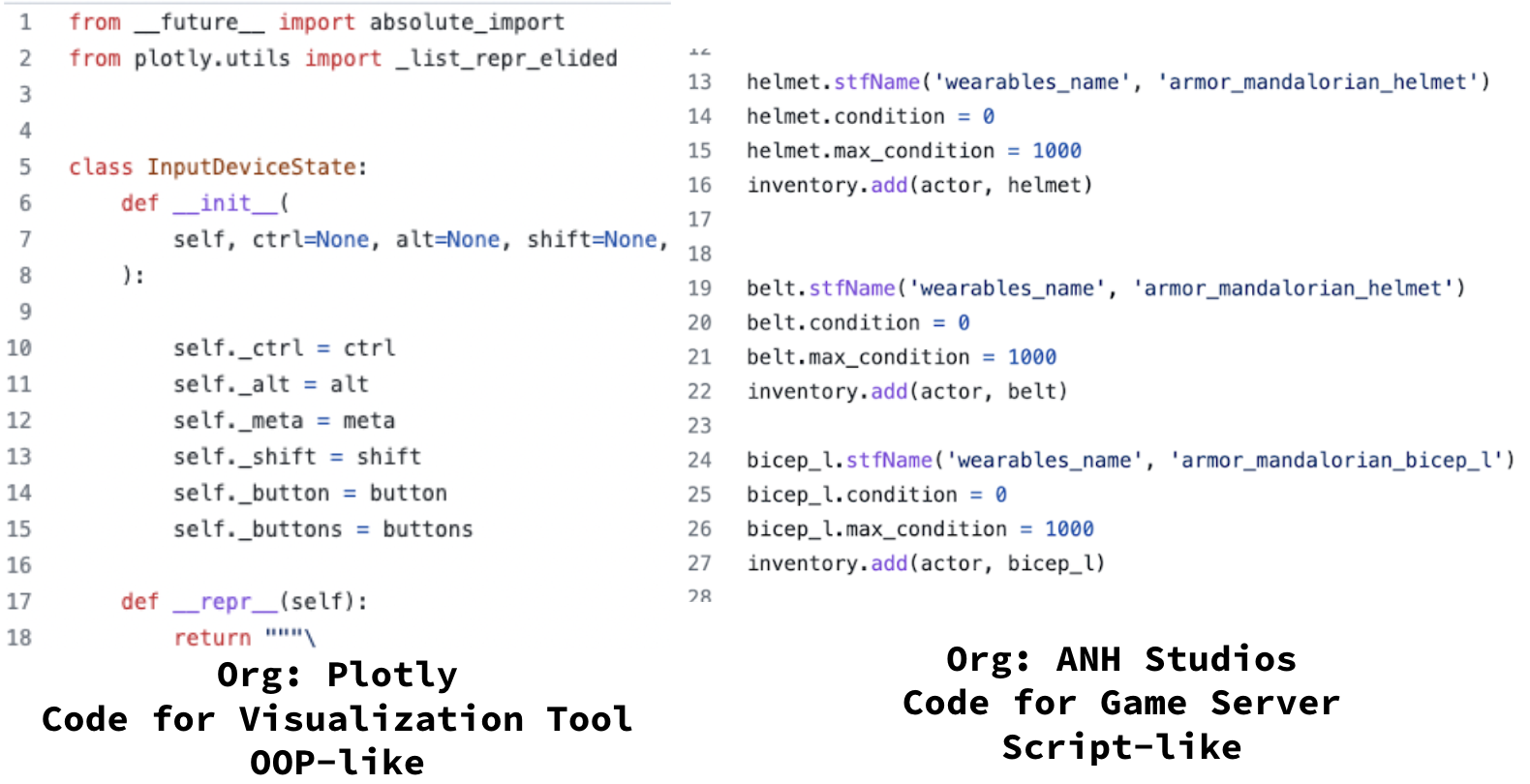}
	\caption{Example of a pair of scripts with distinctive code style: the code from \textit{plotly}, which is used as the component for visualization tools, tends to be object-oriented designed; on the other hand, the code from \textit{anhstudio}, which is used for game server, is having the style of single execution script. They are having difference on class, casing, and docstring usages.}
	\label{style_example_fig}
\end{figure*}

\begin{table*}[ht!]
\begin{center}
\begin{tabular}{| c | c |} 
\hline
\multicolumn{2}{|c|}{Meaning Scale Table} \\
\hline
Scale & Description  \\
\hline
1 & No predicted docstrings have any similar meaning to label docstrings. \\
2 & Some predicted docstrings have similar meaning to labels. \\
3 & More than half predicted docstrings have similar meaning to labels. \\
\hline
\end{tabular}
\caption{Scale Table for Meaning Metric}
\label{meaning_scale_table}
\end{center}
\end{table*}

\begin{figure*}[ht!]
	\centering
	\includegraphics[width=1\textwidth]{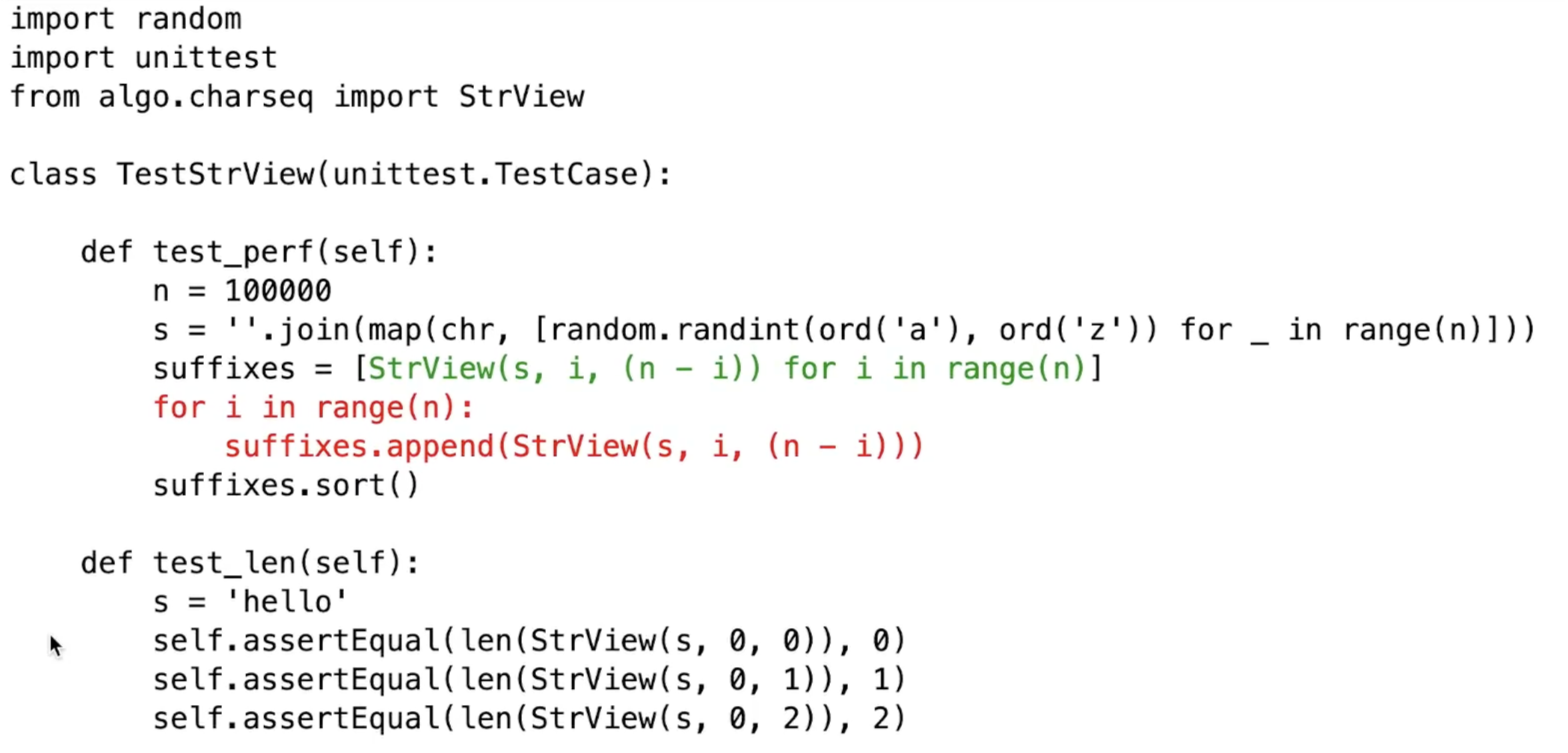}
	\caption{List comprehension transfer examples: the red text is the original text and the green text is the changed part.}
	\label{ind_example_fig}
\end{figure*}

\begin{figure*}[ht!]
	\centering
	\includegraphics[width=1\textwidth]{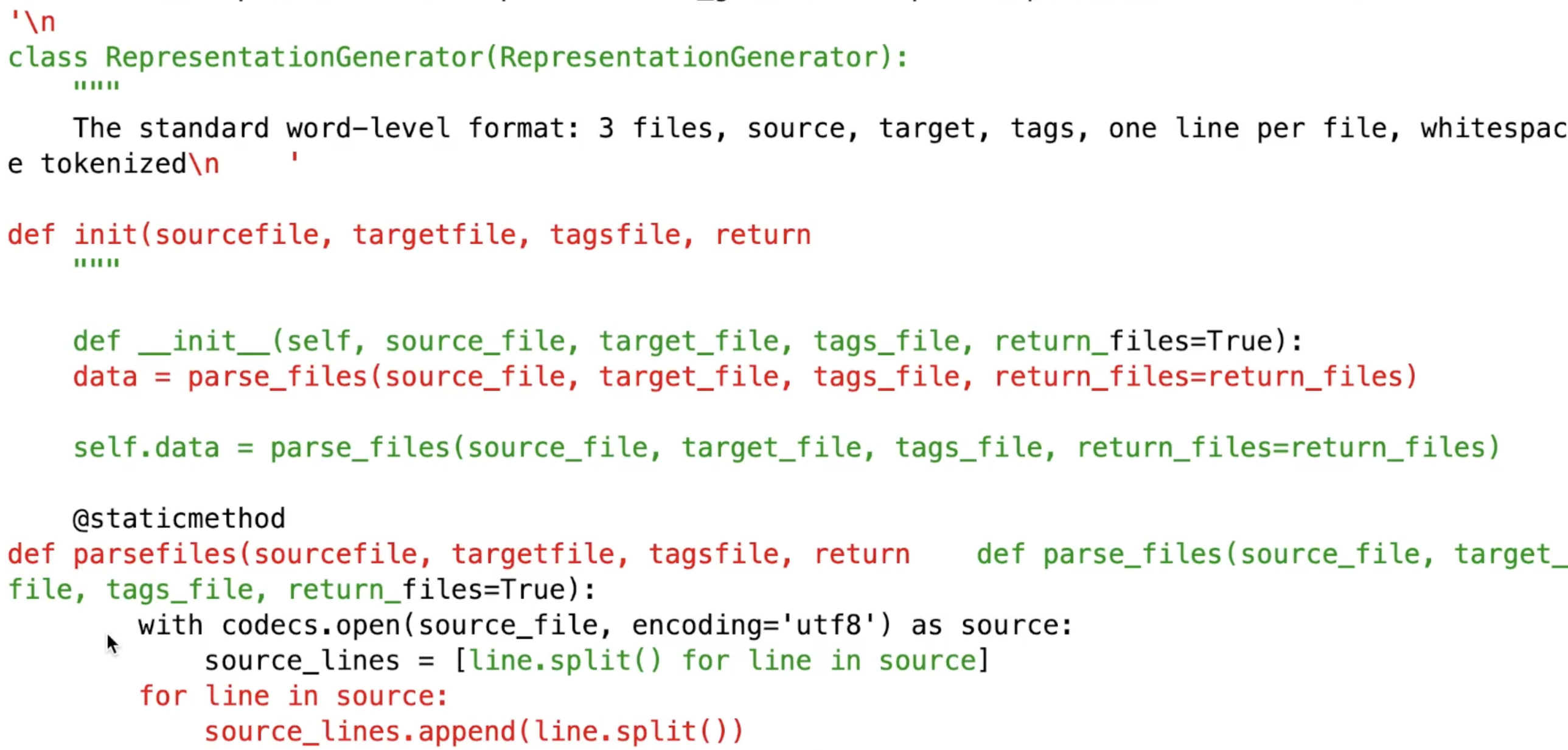}
	\caption{List comprehension + Casing + Class transfer examples: the red text is the original text and the green text is the changed part.}
	\label{mul_example_fig}
\end{figure*}

\end{document}